\documentclass{article}

\usepackage[final]{corl_2023} 

\usepackage{graphicx}
\usepackage{amssymb}
\usepackage{amsbsy}
\usepackage{amsmath}
\usepackage{siunitx}
\usepackage{wrapfig}
\usepackage{subcaption} 
\usepackage{hyperref}
\usepackage{booktabs}
\usepackage{tabularx}
\usepackage{arydshln}
\usepackage[usenames,dvipsnames,svgnames,table]{xcolor}
\usepackage{enumitem,kantlipsum}
\usepackage{array}
\usepackage{pgfplots}

\title{
Learning to See Physical Properties with Active Sensing Motor Policies
}

\author{
  Gabriel B. Margolis \hspace{0.7cm} Xiang Fu \hspace{0.7cm} Yandong Ji \hspace{0.7cm} Pulkit Agrawal\\\
  Improbable AI Lab, Massachusetts Institute of Technology\\
  \protect\url{https://gmargo11.github.io/active-sensing-loco}
}

\begin{document}
\maketitle

\begin{figure}[h]
\vspace{-0.7cm}
\includegraphics[width=\linewidth]{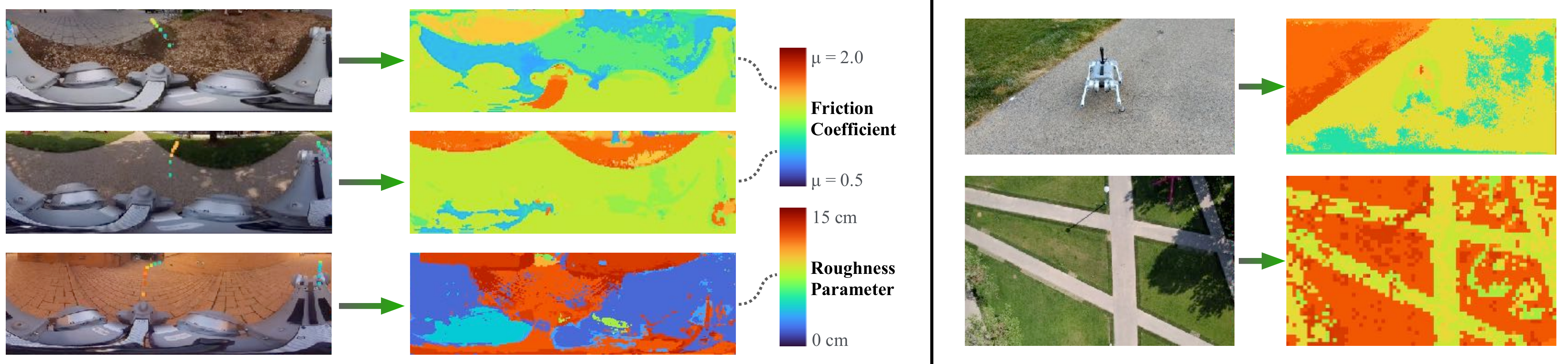}
    \caption{
    \textbf{Learning to see how terrains feel.} \textit{Left}: Sparse training labels acquired from proprioceptive traversal distinguish dirt (top), grass (middle), and stairs (bottom) to supervise dense perception of physical properties. \textit{Right}: Prediction from novel viewpoints facilitates locomotion planning. 
}\label{fig:header}
\end{figure}

\vspace{-0.3cm}
\begin{abstract}
Knowledge of terrain's physical properties inferred from color images can aid in making efficient robotic locomotion plans. However, unlike image classification, it is unintuitive for humans to label image patches with physical properties. Without labeled data, building a vision system that takes as input the observed terrain and predicts physical properties remains challenging. We present a method that overcomes this challenge by self-supervised labeling of images captured by robots during real-world traversal with physical property estimators trained in simulation. To ensure accurate labeling, we introduce \textit{Active Sensing Motor Policies} (ASMP), which are trained to explore locomotion behaviors that increase the accuracy of estimating physical parameters. For instance, the quadruped robot learns to swipe its foot against the ground to estimate the friction coefficient accurately. We show that the visual system trained with a small amount of real-world traversal data accurately predicts physical parameters. The trained system is robust and works even with overhead images captured by a drone despite being trained on data collected by cameras attached to a quadruped robot walking on the ground. 
\end{abstract}


\section{Introduction}

In recent years, legged locomotion controllers have exhibited remarkable stability and control across a wide range of terrains 
such as pavement, grass, sand, ice, slopes, and stairs~\cite{lee2020learning, kumar2021rma, margolis2022rapid, ji2022concurrent, miki2022learning, ji2023dribblebot, agarwal2023legged, choi2023learning}. 
State-of-the-art approaches using sim-to-real learning primarily rely on proprioception and depth sensing to perceive obstacles and terrain~\cite{ miki2022learning, agarwal2023legged, choi2023learning, hoeller2021learning, margolis2021learning, yang2021learning, nahrendra2023dreamwaq, kareer2023vinl, truong2023indoorsim, yang2023neural}. These approaches discard valuable information about the terrain's material properties beyond geometry, such as slip, softness, etc., conveyed by color images. A primary reason for this choice is that sim-to-real transfer has been shown to work with depth images~\cite{miki2022learning,agarwal2023legged,margolis2021learning}, but it remains unclear how well the transfer will work with color or RGB images. To utilize information beyond geometry, some works learn to predict task performance or task-relevant properties (e.g., traversability) from color images using 
data collected in the real world~\cite{hadsell2007online, wellhausen2019should, castro2022does, frey2023fast, yang2023learning}. However, the terrain property predictors learned in prior works are task- or policy-specific, which limits their applicability to new tasks. 

To perceive a multipurpose representation of the terrain, 
we propose predicting the terrain's physical
properties 
(e.g., friction, roughness) that
are invariant to
the policy 
and task. Perceiving the 
\begin{wrapfigure}{r}{0.42\linewidth}
    \centering
    \includegraphics[width=1.0\linewidth]{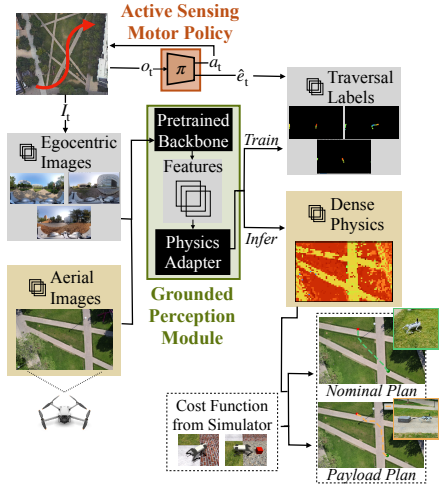}
    \caption{
    \textbf{Active self-supervision.} 
    We propose learning an optimized gait for collecting informative proprioceptive terrain labels that supervise training for a vision module, which can be used for navigation planning with new tasks and views.
    }\label{fig:teaser}
\end{wrapfigure}
physical parameters makes it possible to create a \textit{digital twin} of the terrain in front of the robot and use \textit{simulation} to estimate the cost map for a new task (e.g., dragging a payload) or objective (e.g., preference for speed or energy efficiency). One way to generate the cost map is to collect many rollouts of the policy in simulation and label each location with the value function associated with the new objective. This costmap can be used to plan a trajectory that can be executed in reality. 

The main obstacle in this approach is collecting a dataset of terrain images labeled with associated physical properties.
A natural way of collecting labels is to traverse the terrain and proprioceptively estimate its physical parameters with a neural network supervised by the ground truth labels available in simulation~\cite{ji2022concurrent, miki2022learning,yu2017preparing}. We discovered that the estimates obtained in this way can be imprecise because the locomotion behavior often makes the terrain properties hard to predict. Therefore, unlike prior works in terrain perception that predict the terrain characteristics from passive data~\cite{wellhausen2019should, castro2022does, frey2023fast, yang2023learning}, 
we propose training a specialized data collection policy that directly optimizes for terrain sensing.
This \textit{Active Sensing Motor Policy} (ASMP) learns emergent locomotion behavior, such as dragging the feet on the ground to better estimate friction, and improves the informativeness of its proprioceptive traversals.

We use the improved data obtained through ASMP as self-supervision to learn a visual perception module that predicts terrain material properties (Figure \ref{fig:header}).
The same model can inform efficient plans for nominal locomotion and for dragging objects by considering the impact of terrain properties on traversal cost.
Because the robot is low to the ground, its onboard cameras only provide enough range for local planning. 
Although our model is trained only with data collected by the robot, it can also be evaluated to predict terrain properties using images from various viewpoints.
Therefore, we also consider data from a teamed drone that flies above the legged robot and show that it successfully informs traversal from an extended view of the environment.

\section{Method}
\label{sec:method}

Our approach consists of the following stages, which are also illustrated graphically in Figure \ref{fig:teaser}:

\begin{enumerate}[wide, labelindent=0pt, labelwidth=!]
    \item \textbf{Active Sensing}: We estimate the terrain dynamics parameter, $e_t$, from the proprioceptive sensor history during an initial blind traversal. Our \textit{Active Sensing Motor Policy} (ASMP) crucially provides better-calibrated estimates than the baseline policy.
    In our experiments, the estimated parameter $e_t$ is the ground friction coefficient, the ground roughness magnitude, or both. (Section \ref{sec:active_est})
    \item \textbf{Self-Supervised Vision Learning}: Using labels of $e_t$ recorded from the real-world traversal of the robot, 
    we learn a function, $\hat{e} = f(\mathbf{I})$, that predicts the per-pixel value of $e_t$ for a given image $\mathbf{I}$. The labels for training are only available at the pixels corresponding to the places the robot traversed, but the resulting model can be queried to predict the terrain parameter at any pixel.
    (Section \ref{sec:dino_ground})
    
    \item \textbf{Cost Function Learning}: To inform planning, we learn cost functions that relate the terrain dynamics to various performance metrics. First, we create terrains with a range of $e_t$ values in simulation. 
    Then, we perform rollouts in simulation to measure a
    cost function
    $C({e}_k)$ that relates dynamics parameters to performance. We learn a separate cost function for each task. 
    (Section \ref{sec:affordance_learning})
    \item \textbf{Dynamics-Aware Path Planning}: Combining (2-3), we compute cost maps directly from color images and use them for path planning. (Section \ref{sec:integrated_planning})
\end{enumerate}

\begin{figure}
    \centering
    \includegraphics[width=\linewidth]{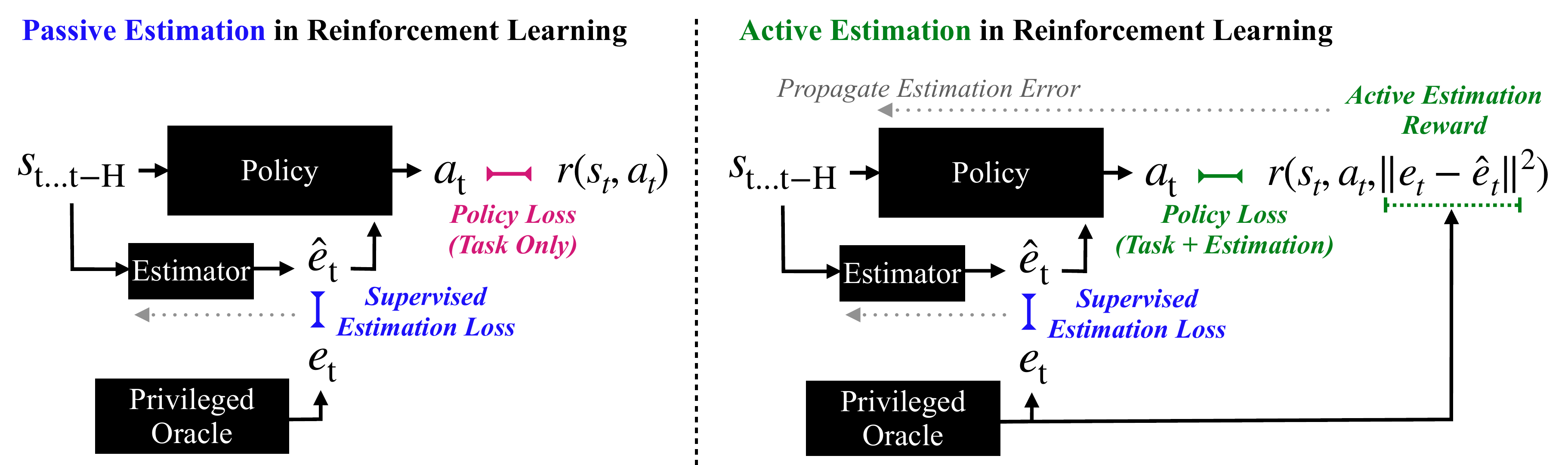}
    \caption{\textbf{Active Sensing Motor Policies optimize for estimation.}
    When training in simulation, an oracle can provide privileged state information $e_t$ that cannot be directly measured on the real robot, such as the present roughness and friction of the terrain. Prior works learn an estimator network to make predictions $\hat{e}_t$ from the history of sensor readings that are available on the real robot $s_{t...t-H}$ (left). We propose additionally optimizing the policy for estimation (right).
    This incentivizes information-gathering behaviors, like intentionally swiping the robot's foot during legged locomotion to estimate the terrain properties more accurately.
    }
    \label{fig:active_sensing_arch}
\end{figure}

\subsection{Active Sensing Motor Policies: Learning Whole-Body Active Estimation} \label{sec:active_est}

In learning control policies under partial observations, it is commonplace to train with an implicit~\cite{lee2020learning, kumar2021rma, margolis2022rapid, miki2022learning, choi2023learning} or explicit~\cite{ji2022concurrent, margolis2022walktheseways} incentive to form representations within the policy network that correspond to the unobserved dynamics parameters. Consider the concurrent state estimation framework of \citet{ji2022concurrent}, where a state estimation network is trained simultaneously with the policy network to predict the unobserved parameters. The predictions of the state estimation network are concatenated with the rest of the observation to construct the policy network input. This approach optimizes a two-part loss consisting of the standard policy objective and the state estimation error: $L(\theta, \theta') = \hat{\mathbb{E}}_t[ \log \pi_\theta(a_t|s_t, \hat{e}_t)\hat{A}_t  ] +  \lVert e_t - \hat{e}_{\theta'}(s_t)\rVert^2\textrm{.}$ This has been empirically shown to yield better policy performance in environments with randomized dynamics or partial observations~\cite{ji2022concurrent, yu2017preparing}. 

In the formulation above, the estimation error is used to update the state estimator weights $\theta'$, but not the policy weights $\theta$. This does not incentivize the \textit{policy} to adjust its actions to improve estimation performance beyond what is required for control. 
Typically, this is no problem because it allows the policy to maximize its performance at the current control task. 
However, our end goal is to use the output of the state estimator to train a visual perception module that may be reused with other controllers and tasks. To support this, the labels should be as accurate as possible even when that is not necessary for control.  
To obtain the most accurate perception module, we would like a mechanism to improve the state estimate quality of the proprioceptive data collection policy as much as possible by adapting the policy's behavior.
To this end, we propose \textit{Active Sensing Motor Policies} in which the policy $\pi^\mathrm{est}$ is trained with an additional \textit{estimation reward}: $r_\mathrm{est} = c\cdot\exp{(\lVert e - \hat{e}\rVert^2)}$. Figure \ref{fig:active_sensing_arch} illustrates the policy architecture.
In practice, we observe that an Active Sensing Motor Policy that is rewarded for estimating the ground friction coefficient slides one foot along the ground or swipes it vigorously to improve the friction coefficient observability in the state history.

\subsection{Grounding Visual Features in Physics from Real-world Experience} \label{sec:dino_ground}

We collect paired proprioceptive and vision data from the state estimation policy in the real world in order to learn about the relationship between visual appearance and terrain physics. Specifically, we collect data of the form $(\mathbf{I}, \hat{e}, \mathbf{x})_t$ where $\mathbf{I}$ is a camera image, $\hat{e}$ are the estimated dynamics parameters and $\mathbf{x}$ is the position and orientation of the robot in a fixed reference frame. We obtain $\mathbf{x}$ by training an additional 2D output of the final MLP layer
in our learned state estimator to predict the displacement in the ground plane of the base from its location at the previous timestep, $\Delta \mathbf{x}$, 
and then integrate the estimated displacements. The integrated estimates $\mathbf{x}$ will drift over time, but we will only rely on them over a short time window.  
This alleviates the need for a separate odometry algorithm to estimate the robot's state.

Using the camera intrinsic and extrinsic transform, we project the relative positions of the robot in the past and future 
into each camera image frame. We restrict the positions to those between \SI{1}{\meter} and \SI{5}{\meter} from the robot along the traversal path so that they are neither too far away to see nor so close as to be obstructed from view by the robot's own body. We label each of the projected robot positions with the estimated dynamics parameters $\hat{e}$ that the robot felt when it walked there. This yields a corresponding label image $\mathbf{I}^e_t$ for each color image $\mathbf{I}$ where the traversed pixels are labeled with their measured dynamics.

For each color frame $\mathbf{I}_t$, we use the pretrained convolutional backbone \cite{yu2023convolutions} to compute a dense feature map. 
Similar to the procedure that \citet{oquab2023dinov2} used for depth estimation, we discretize the labels $\hat{e}_t$ into $20$ bins and train a single linear layer with cross-entropy loss where the inputs are the features of one patch and the outputs are the logits of the patch's $\hat{e}_t$ label from proprioception.

\subsection{Cost Function Learning: Connecting Physics Parameters to Affordances} \label{sec:affordance_learning}

\begin{wrapfigure}{r}{0.5\textwidth}
    \vspace{-0.5cm}
    \includegraphics[width=0.5\textwidth]{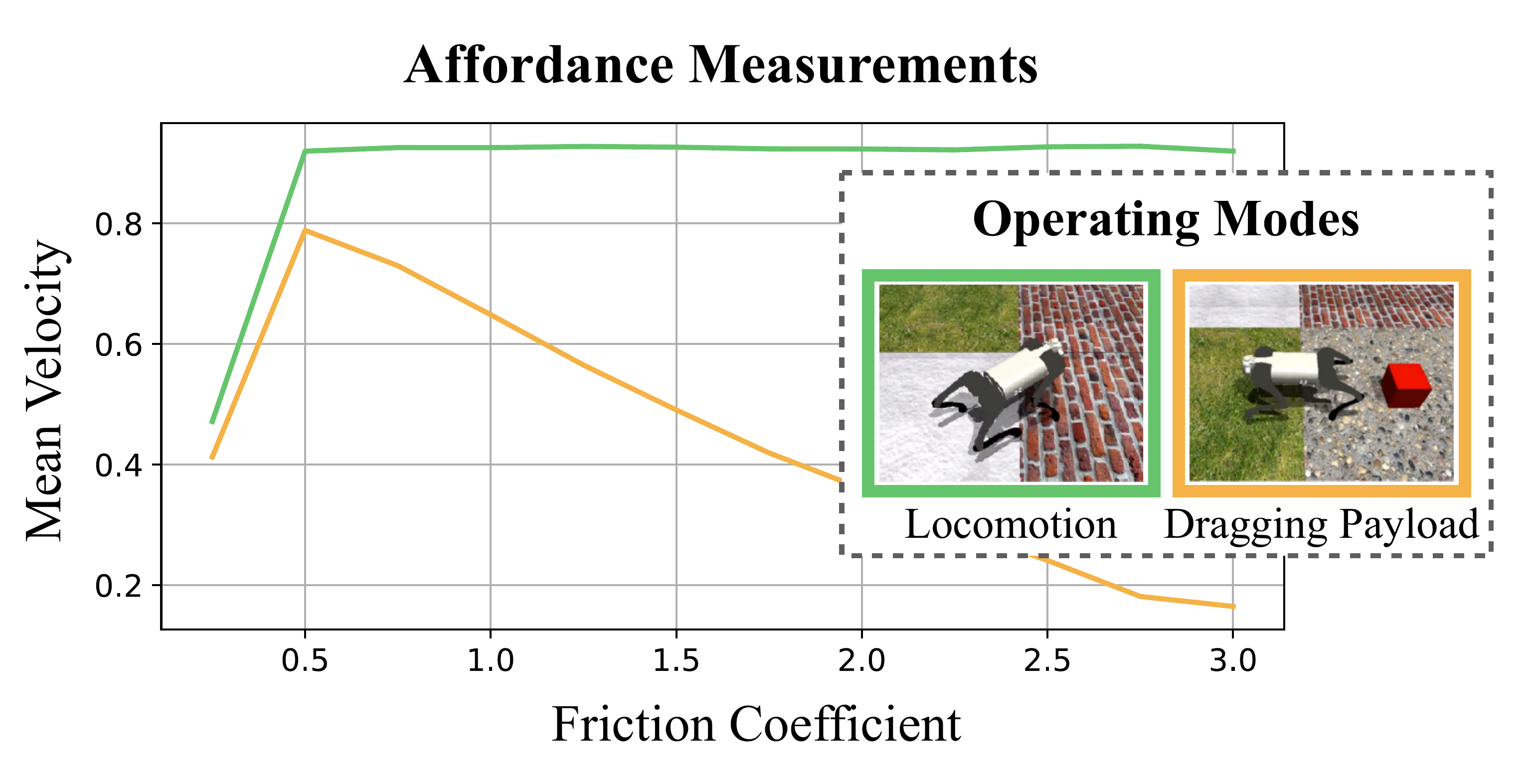}
    \caption{\textbf{Locomotion affordances.} We measure the dependence of locomotion performance (\SI{1}{\meter/\second}) on terrain friction in two different operating modes. In free locomotion, the controller maintains the target velocity across a range of friction coefficients, except for the lowest friction. In contrast, when dragging a weighted box, the robot slows down as the terrain friction increases.
    }
    \label{fig:affordance_meas}
\end{wrapfigure}

The impact of terrain properties on robot performance is task-dependent: for example, a robot dragging 
an object may face distinct constraints that inhibit its traversal on some terrains, compared
to a robot without any payload. To use our vision module for planning, we must establish a mapping between terrain properties and robot performance for each task. We propose a simple procedure for extracting a task cost function from simulated data to demonstrate that our perception module can be useful in planning for multiple tasks, which we refer to as ``operating modes". We sample simulated terrains with a variety of terrain properties $e_t$ and command a locomotion policy from prior work~\cite{margolis2022walktheseways} to walk forward at \SI{1}{\meter/\second}. We record the actual resulting velocity achieved on each terrain. We evaluate the mean realized velocity for multiple operating modes: (1) locomotion, (2) payload dragging. We construct a cost function for each operating mode as the average time spent traversing one meter of a given terrain. Minimizing this cost function during path planning will yield an estimated shortest-time path. While we focus on time-optimal payload dragging as an example, (1, 2) could be any combination of task and metric as long as their relation to terrain properties can be evaluated in simulation. 

\subsection{Integrated Dynamics-Aware Path Planning from Vision} \label{sec:integrated_planning}

Our perception module (Section \ref{sec:dino_ground}) runs in real-time (\SI{2}{\hertz}) using onboard compute. Although it was trained using images from a 360-degree camera, 
the resulting pixel-wise friction estimator can be evaluated in images from other cameras including the robot's onboard fisheye camera and an overhead drone. This is useful because the perception module can remain useful when deployed on a new robot or evaluated from a new viewpoint.

One possible scenario for carrying ground objects across a long distance is that of a drone-quadruped team.
In this case, we can directly evaluate our grounded vision module in overhead images to obtain a pixel-wise friction mask. Then, considering the robot's operating mode, we compute the cost associated with each pixel using the corresponding cost function determined from simulation (Section \ref{sec:affordance_learning}). Given this overhead cost map, we use the A$^*$ search algorithm \cite{hart1968formal} to compute the minimum cost traversal path for the current operating state.

\subsection{System Setup}
\label{sec:materials}

\textbf{Robot}: We use the Unitree Go1 robot, a \SI{12}{}-motor quadruped robot standing \SI{40}{\centi\meter} tall. It has an NVIDIA Jetson Xavier NX processor, which runs the control policy and the vision module. For payload dragging experiments, the robot's body is connected to an empty suitcase using a rope.

\textbf{360 Camera}: We use an Insta360 X3 360 action camera mounted on the robot to collect images for training the perception module. This camera provides a \SI{360}{\degree} field of view. Before the image data is used for training, we use the Insta360 app to perform image stabilization, which takes about two minutes for data collected from a ten-minute run.

\textbf{Training Compute}: We perform policy training, video postprocessing, and vision model training on a desktop computer equipped with an NVIDIA RTX 2080 GPU. 

\textbf{Drone Camera}: For planning from overhead images, we record terrain videos using a DJI Mini 3, a consumer camera drone.


\section{Results}
\label{sec:result}

\begin{figure}[t!]
    \begin{subfigure}[b]{0.5\textwidth}
        \includegraphics[width=0.48\textwidth]{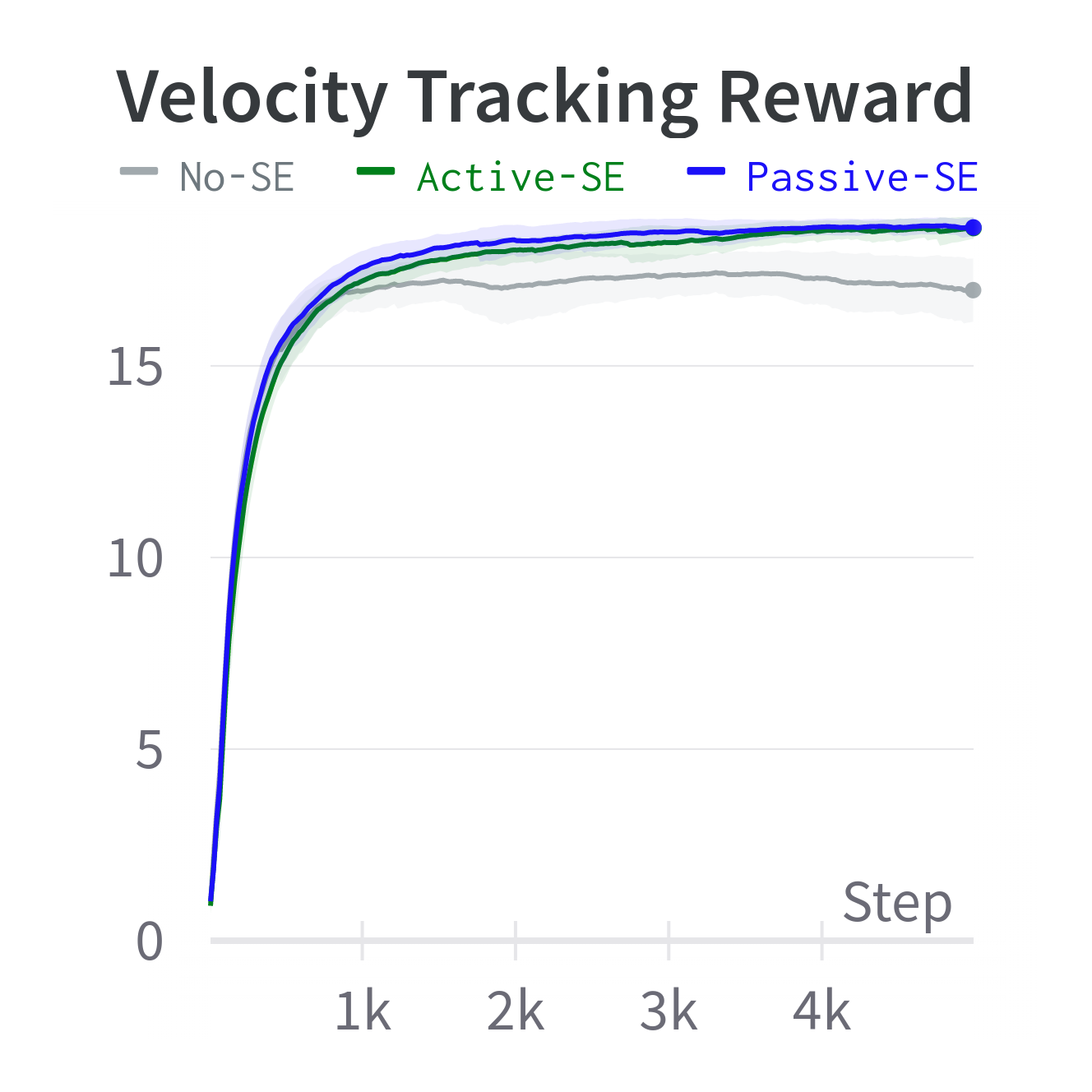}
        \includegraphics[width=0.48\textwidth]{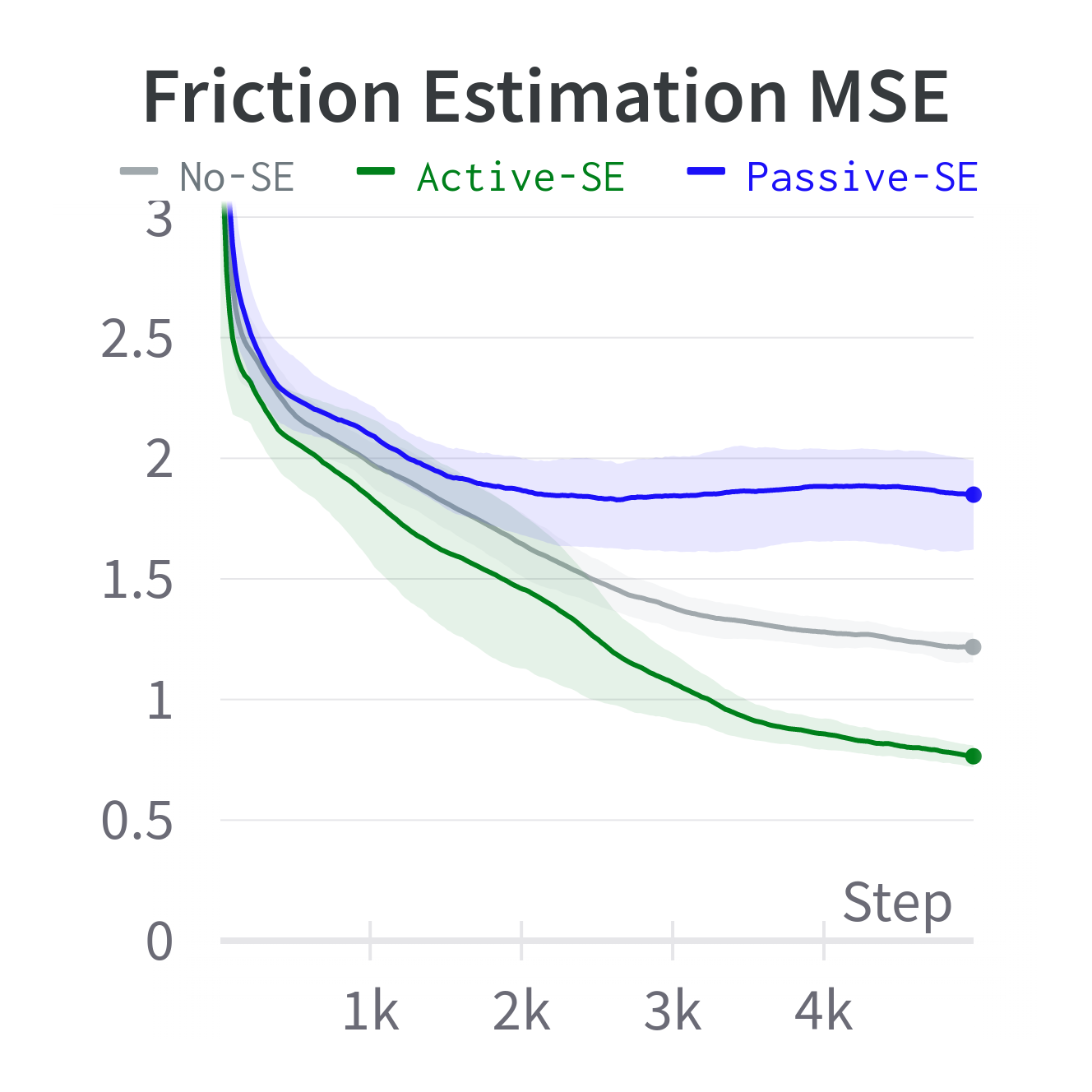}
        \caption{Performance and estimate quality during training.}
        \label{fig:nose_compare}
    \end{subfigure}
    \begin{subfigure}[b]{0.5\textwidth}
        \includegraphics[width=\textwidth]{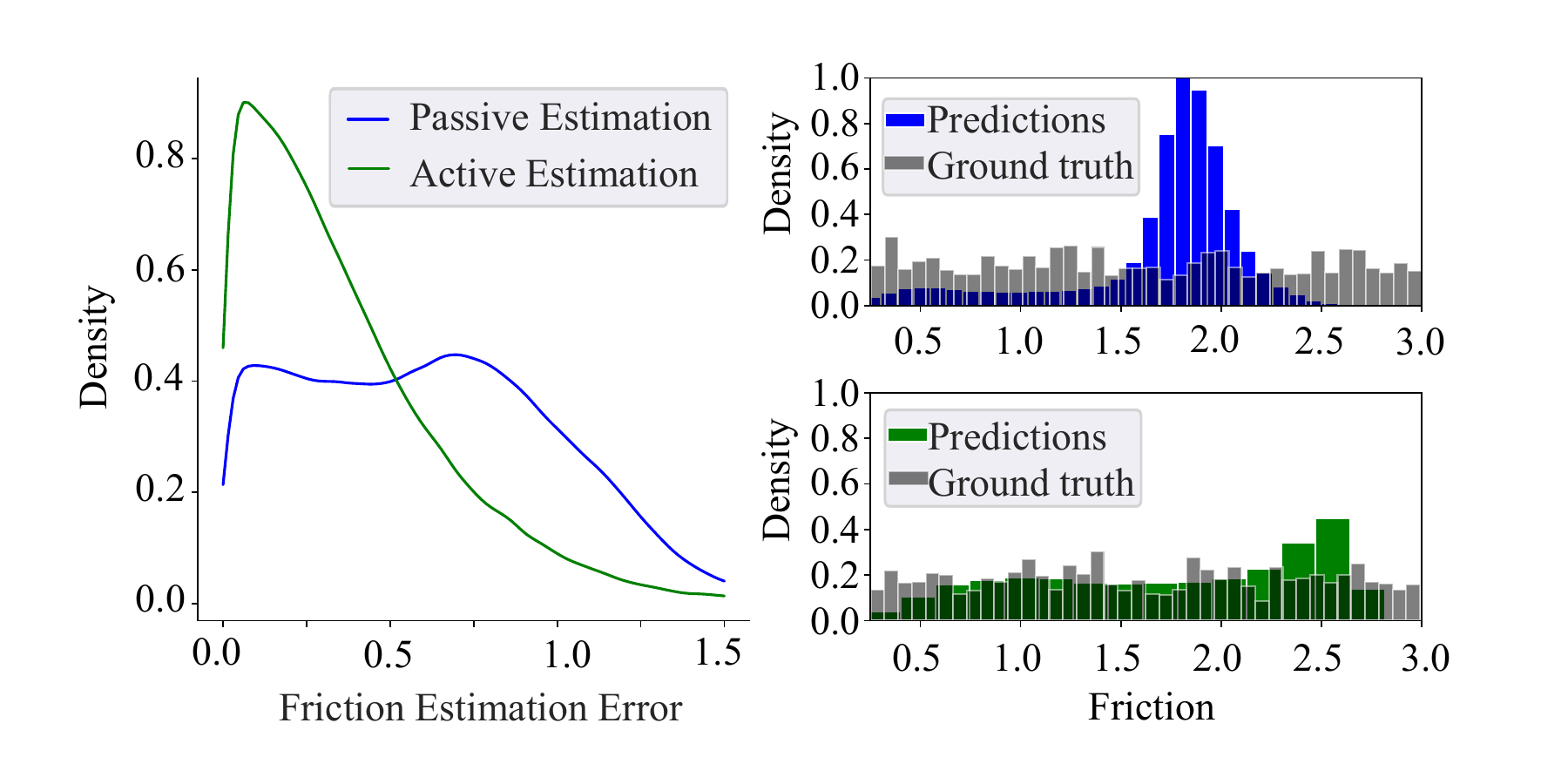}
        \caption{Distribution of friction estimates at convergence.}
        \label{fig:ase_histcomp}
    \end{subfigure}
    \caption{\textbf{Learning active estimation.} Active Sensing Motor Policies (\texttt{Active-SE}) automatically learn motor skills (e.g. dragging the feet) that improve observability of the environment properties.
    }
    \label{fig:active_sensing_perf}
\end{figure}

\subsection{Interaction among Estimation, Adaptation, and Performance} 

\textit{Observing supervised internal state estimates improves proprioceptive locomotion.} Affirming the results of \citet{ji2022concurrent}, we train a state estimation network using supervised learning to predict privileged information (the ground friction coefficient and terrain roughness parameter) from the history of sensory observations. When the policy is allowed to observe the output of this state estimation network (\texttt{Passive-SE}), the policy training is more stable and results in a more performant final policy than when the state estimate is not observed (\texttt{No-SE}) (Figure \ref{fig:active_sensing_perf}). 

\textit{Observing passive state estimates can degrade the state observability.} We compute the error distribution of the learned state estimator in \texttt{Passive-SE} and \texttt{No-SE} policies (Figure \ref{fig:active_sensing_perf}). It may be surprising that the friction estimation error of the more-performant \texttt{Passive-SE} policy is \textit{higher} than that of the less-performant \texttt{No-SE} policy. We suggest an explanation for this: Supposing some irreducible sensor noise, two terrains of different frictions will only be distinguishable if they make the robot slip in sufficiently different ways. However, a control policy with a better adaptive facility is more likely to avoid slipping across a wide range of ground frictions. Because slip occurs less frequently in the more adaptive policy, the observability of the ground friction coefficient degrades.

\textit{Our method, ASMP, produces the best privileged state observability.}  We train an active sensing motor policy (\texttt{Active-SE}) to intentionally measure the friction as described in Section \ref{sec:active_est}. (The full reward function for each policy we trained is provided in the appendix.) We find that the \texttt{Active-SE} policy provides the most accurate friction estimates among the three architectures  (Figure \ref{fig:active_sensing_perf}). Therefore, as we will further show, it is the superior policy for supervising a task-agnostic physical grounding for vision.

\definecolor{matteGrass}{HTML}{68855C}
\definecolor{mattePavement1}{HTML}{707070}
\definecolor{matteDirt}{HTML}{8B6F57}
\definecolor{matteGravel}{HTML}{8A8275}
\definecolor{colorMeasured}{HTML}{000000}
\definecolor{colorASMP}{HTML}{00801B}
\definecolor{colorPassive}{HTML}{2922F8}
\definecolor{colorVisionTrain}{HTML}{DA702C}
\definecolor{colorVisionTest}{HTML}{D0A215}

\begin{figure}
\centering
\small
\begin{tikzpicture}
    \begin{axis}[
        width=12cm,
        height=4.5cm,
        enlarge x limits=0.15,
        enlarge y limits=0.2,
        legend style={
            at={(0.5,1.25)},
            anchor=north,
            legend columns=5,
        },
        ylabel={$\mu$ (Friction Coefficient)},
        xtick={1,2,3,4},
        xticklabels={Grass, Pavement, Dirt, Gravel},
        cycle list={{colorMeasured},{colorASMP},{colorPassive},{colorVisionTrain},{colorVisionTest}}, 
        error bars/error bar style={line width=0.75pt},  
        every mark/.append style={mark size=3pt},
        ytick align=outside,
        xtick align=outside,
        x axis line style={-},
        y axis line style={-},
        xtick pos=left,
        ytick pos=left,
        xtick style={color=black},
        ytick style={color=black},
        tickwidth=0pt,
        xmajorgrids=true,
        ymajorgrids=true,
    ]
    
    \pgfmathsetmacro{\offset}{0.1}

    
    \addplot+[only marks, mark=square*, mark options={fill=colorMeasured}, error bars/.cd, y dir=both,y explicit]
        coordinates {(1-2*\offset,1.45) +- (0,0)};
    \addplot+[only marks, mark=o, mark options={fill=colorASMP}, error bars/.cd, y dir=both,y explicit]
        coordinates {(1-1*\offset,1.92) +- (0,0.28)};
    \addplot+[only marks, mark=triangle*, mark options={fill=colorPassive}, error bars/.cd, y dir=both,y explicit]
        coordinates {(1-0*\offset,2.58) +- (0,0.14)};
    \addplot+[only marks, mark=oplus, mark options={fill=colorVisionTrain}, error bars/.cd, y dir=both,y explicit]
        coordinates {(1+1*\offset,1.89) +- (0,0.18)};
    \addplot+[only marks, mark=diamond*, mark options={fill=colorVisionTest}, error bars/.cd, y dir=both,y explicit]
        coordinates {(1+2*\offset,1.80) +- (0,0.26)};

    \addplot+[only marks, mark=square*, mark options={fill=colorMeasured}, error bars/.cd, y dir=both,y explicit]
        coordinates {(2-2*\offset,0.89) +- (0,0)};
    \addplot+[only marks, mark=o, mark options={fill=colorASMP}, error bars/.cd, y dir=both,y explicit]
        coordinates {(2-1*\offset,1.35) +- (0,0.29)};
    \addplot+[only marks, mark=triangle*, mark options={fill=colorPassive}, error bars/.cd, y dir=both,y explicit]
        coordinates {(2-0*\offset,2.42) +- (0,0.13)};
    \addplot+[only marks, mark=oplus, mark options={fill=colorVisionTrain}, error bars/.cd, y dir=both,y explicit]
        coordinates {(2+1*\offset,1.32) +- (0,0.22)};
    \addplot+[only marks, mark=diamond*, mark options={fill=colorVisionTest}, error bars/.cd, y dir=both,y explicit]
        coordinates {(2+2*\offset,1.35) +- (0,0.22)};

    \addplot+[only marks, mark=square*, mark options={fill=colorMeasured}, error bars/.cd, y dir=both,y explicit]
        coordinates {(3-2*\offset,0.63) +- (0,0)};
    \addplot+[only marks, mark=o, mark options={colorASMP}, error bars/.cd, y dir=both,y explicit]
        coordinates {(3-1*\offset,0.90) +- (0,0.23)};
    \addplot+[only marks, mark=triangle*, mark options={fill=colorPassive}, error bars/.cd, y dir=both,y explicit]
        coordinates {(3-0*\offset,2.12) +- (0,0.26)};
    \addplot+[only marks, mark=oplus, mark options={fill=colorVisionTrain}, error bars/.cd, y dir=both,y explicit]
        coordinates {(3+1*\offset,0.90) +- (0,0.23)};
    \addplot+[only marks, mark=diamond*, mark options={fill=colorVisionTest}, error bars/.cd, y dir=both,y explicit]
        coordinates {(3+2*\offset,1.09) +- (0,0.36)};

    \addplot+[only marks, mark=square*, mark options={fill=colorMeasured}, error bars/.cd, y dir=both,y explicit]
        coordinates {(4-2*\offset,0.74) +- (0,0)};
    \addplot+[only marks, mark=o, mark options={fill=colorASMP}, error bars/.cd, y dir=both,y explicit]
        coordinates {(4-1*\offset,0.90) +- (0,0.39)};
    \addplot+[only marks, mark=triangle*, mark options={fill=colorPassive}, error bars/.cd, y dir=both,y explicit]
        coordinates {(4-0*\offset,1.75) +- (0,0.47)};
    \addplot+[only marks, mark=oplus, mark options={fill=colorVisionTrain}, error bars/.cd, y dir=both,y explicit]
        coordinates {(4+1*\offset,0.81) +- (0,0.26)};
    \addplot+[only marks, mark=diamond*, mark options={fill=colorVisionTest}, error bars/.cd, y dir=both,y explicit]
        coordinates {(4+2*\offset,1.12) +- (0,0.39)};
    
    \addlegendimage{mark=square*,mark size=1pt,mark options={color=colorMeasured},color=colorMeasured}
    \addlegendentry{\texttt{Measured}}
    \addlegendimage{mark=o,mark size=1pt,mark options={color=colorASMP},color=colorASMP}
    \addlegendentry{\texttt{ASMP (Ours)}}
    \addlegendimage{mark=triangle*,mark size=1pt,mark options={color=colorPassive},color=colorPassive}
    \addlegendentry{\texttt{Passive (Baseline)}}
    \addlegendimage{mark=oplus,mark size=1pt,mark options={color=colorVisionTrain},color=colorVisionTrain}
    \addlegendentry{\texttt{Vision (Train)}}
    \addlegendimage{mark=diamond*,mark size=1pt,mark options={colorVisionTest},color=colorVisionTest}
    \addlegendentry{\texttt{Vision (Test)}}
    
    \end{axis}
\end{tikzpicture}

\caption{\textbf{Real-world friction sensing performance with proprioception and vision.} \texttt{Measured} values are directly measured by a dynamometer. The predictions from our proposed \texttt{ASMP (Ours)} agree better with the dynamometer measurements than the baseline \texttt{Passive (Baseline)}. \texttt{Vision (Train)} shows the generalization of visual prediction to un-traversed patches in the training images from the onboard camera; (\texttt{Vision (Test)}) shows the generalization to unseen patches and viewpoints by evaluating on drone footage. We use manual segmentation maps (Appendix Figure \ref{fig:segmentation_example}) to match pixel predictions to terrains. Error bars indicate one standard deviation.
}
\label{fig:friction_eval}
\end{figure}
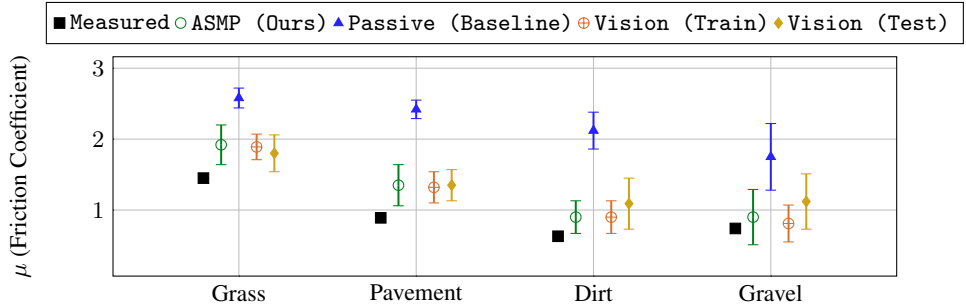

\subsection{Learning to See Physical Properties} 

\textbf{Real-world Evaluation.} We collect fifteen minutes of real-world traversal data spanning diverse terrains: grass, gravel, dirt, pavement, and stairs. Following the procedure in Section \ref{sec:dino_ground}, we project the traversed points into the corresponding camera images and train a linear head on top of a convolutional backbone pretrained for segmentation~\cite{yu2023convolutions} to predict the terrain friction and roughness estimate for each traversed patch. To evaluate estimation performance in the real world, we manually label image segments in a subset of train and test images containing grass, pavement, dirt, or gravel and compute the distribution of proprioceptive and visual friction predictions for each (Figure \ref{fig:friction_eval}). We measure a ground truth friction value for each terrain using a dynamometer by measuring the weight of a payload made of robot foot material and its drag force. The proprioceptive estimates from ASMP are much closer to the dynamometer measurements than the estimates from the passive baseline. They do not match perfectly, suggesting a small but measurable sim-to-real gap in the robot dynamics or terrain modeling. They agree with the dynamometer measurements on the ordering of terrains from most to least slippery. The grounded vision module is close to the distribution of proprioceptive estimates for both train and test images, with increased variance in test images. 

\subsection{Integrated Planning} 

\textbf{Cost Function Evaluation.} We define a cost metric for the locomotion policy from \cite{margolis2022walktheseways} as the distance traveled per second when commanded with a speed of \SI{1.0}{\meter/\second}. We evaluate this metric in simulation by averaging the performance of 50 agents simulated in parallel for \SI{20}{\second} on terrains of different friction coefficients ranging from a lower limit of $\mu=0.25$ to an upper limit of $\mu=3.0$. This procedure is performed once with the robot in nominal locomotion and again with the robot dragging a \SI{1.0}{\kilo\gram} payload. Figure \ref{fig:affordance_meas} shows the measured result; both tasks yield poor performance on extremely slippery terrain, but on higher terrains, the robot dragging a payload slows down while the free-moving robot adapts to maintain velocity. Knowledge of the ground's physical properties motivates a difference in high-level navigation decisions between the two tasks.

\begin{figure}
    \centering
    \begin{minipage}[c]{0.5\textwidth}
        \vspace{-0.3cm}
        \raisebox{-0.5\height}{\makebox[2em][l]{(a)}}\raisebox{-0.5\height}{\includegraphics[width=\linewidth]{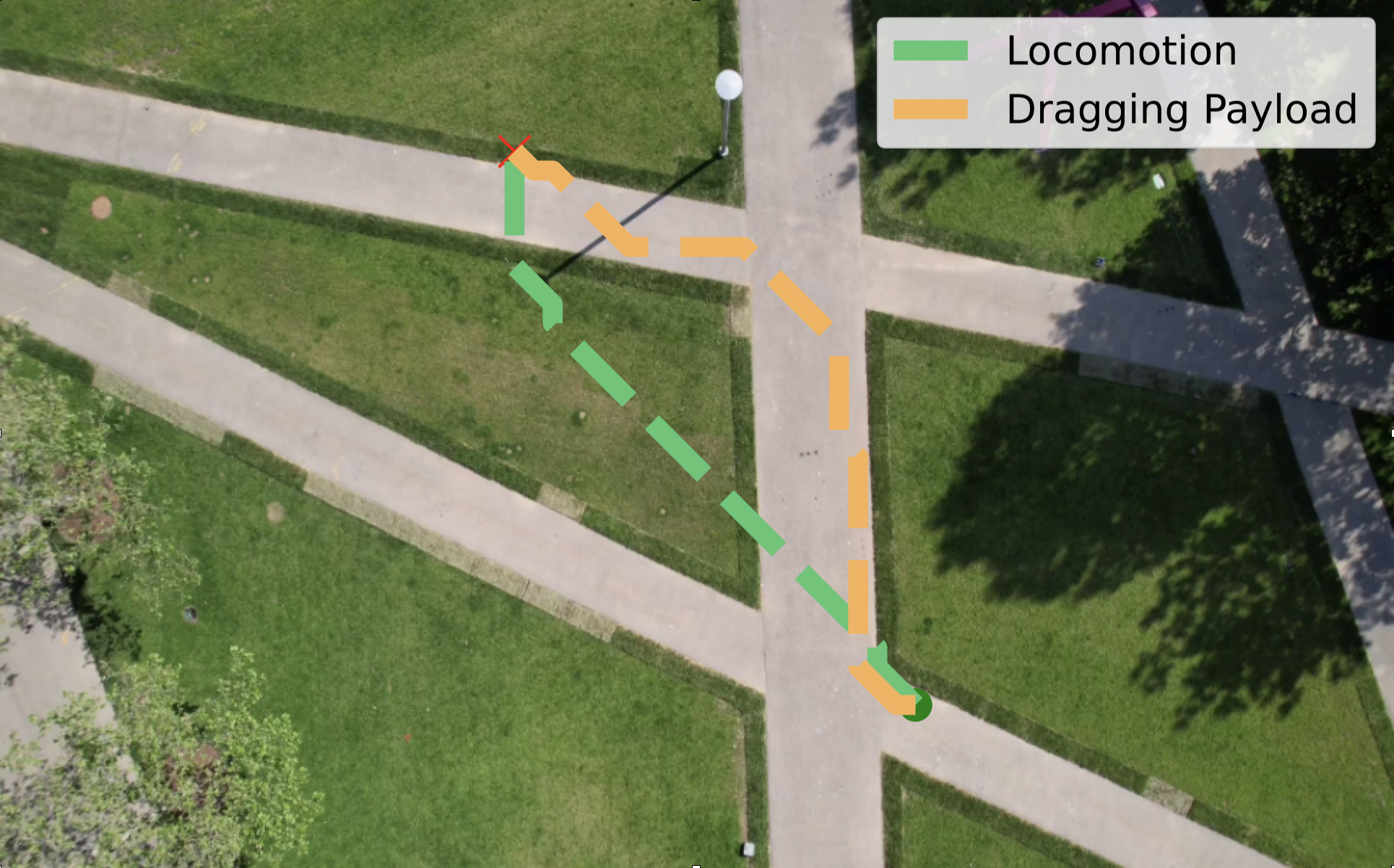}}
    \end{minipage}
    \hspace{0.8cm}
    \begin{minipage}[c]{0.24\textwidth}
        \raisebox{-0.5\height}{\includegraphics[width=\linewidth]{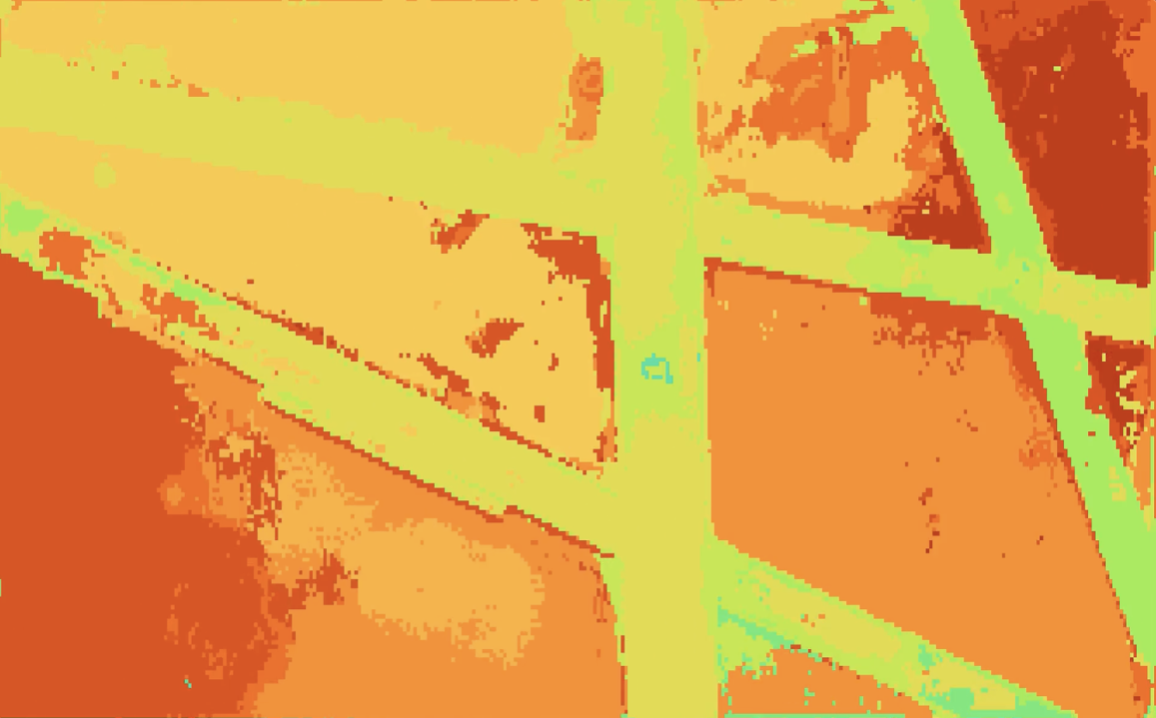}}\raisebox{-0.5\height}{\makebox[2em][r]{(b)}}
        \vspace{1em}
        \raisebox{-0.5\height}{\includegraphics[width=\linewidth]{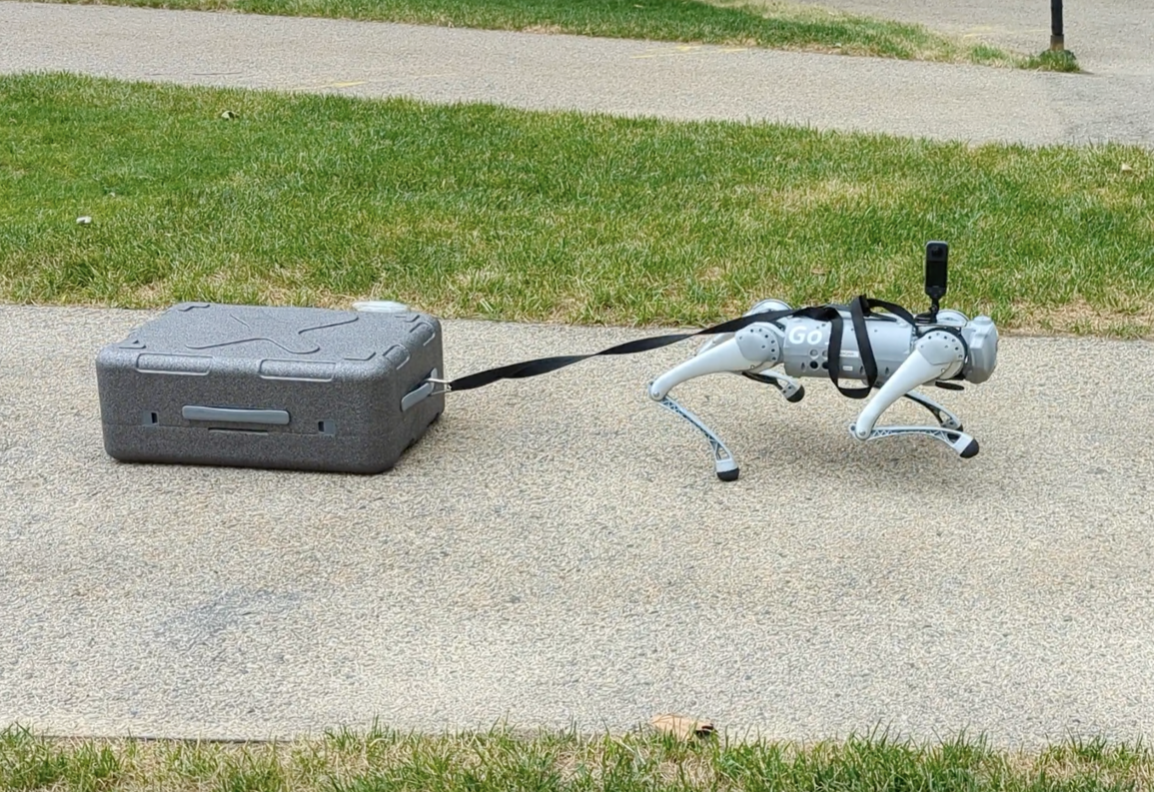}}\raisebox{-0.5\height}{\makebox[2em][r]{(c)}}
    \end{minipage}
    \newline
    \begin{minipage}[c]{\textwidth}
    \footnotesize
    \renewcommand{\arraystretch}{1.5} 
\begin{tabularx}{1.0\textwidth}{|X|X|X|X|}
    \hline
    \textbf{Operating Mode} & \textbf{Metric} & \textbf{Cross Grass} & \textbf{Stay on Sidewalk} \\ 
    \hline
    Dragging Payload & Time (\SI{}{\second}) & $48\pm1$ & $45\pm1$ \\
    Locomotion & Time (\SI{}{\second}) & $23\pm1$ & $26\pm0$ \\
    \hline
\end{tabularx}
    \end{minipage}
    \caption{\textbf{Path planning in overhead images.} (a) We use the learned vision module to plan navigation in overhead images of terrain. (b) The vision module is only trained using first-person views from the robot but can infer the terrain friction with a different camera model and viewing pose. (c) We teleoperate the robot across both planned paths in each locomotion mode. The preference among paths in the real world matches the planning result from our pipeline.
    }
    \label{fig:real_world_plan_execution}
\end{figure}

\textbf{Path Planning and Execution.} We plan paths for locomotion and payload dragging and execute them via teleoperation to evaluate whether the predicted preferences hold true in the real world. We fly a drone over the same environment where the vision model was trained and choose a bird's-eye-view image that includes grass and pavement. We estimate the friction of each pixel and from this we compute the associated cost for locomotion and payload dragging. Then we use A$^*$ search to compute optimal paths. The optimized paths and traversal result are shown in Figure \ref{fig:real_world_plan_execution}. In agreement with the planning result, it is preferred to remain on the sidewalk while dragging the payload and cut directly across the grass when in free locomotion.


\section{Related Work}
\label{sec:conclusion}

Self-supervised traversability estimation has been studied previously for the navigation of wheeled and legged robots. Some works have focused on the direct estimation of a traversability metric, a scalar value quantifying the cost of traversing a particular terrain \cite{castro2022does, frey2023fast, fu2022coupling}. These approaches are specialized to the robot's traversal capability at the time of data collection, implying that a change in the policy or task may necessitate repeated data collection to train a new vision module.

Other works have demonstrated self-supervised terrain segmentation from proprioceptive data~\cite{wellhausen2019should, wu2016integrated, lysakowski2022unsupervised}. 
\citet{wu2016integrated} demonstrated that proprioceptive data from a C-shaped leg equipped with tactile sensors may be sufficient to classify different terrains. \citet{wellhausen2019should} took supervision from the dominant features of a six-axis force-torque foot sensor during traversal and trained a model to densely predict a ground reaction score from color images to be used for planning. 
\citet{lysakowski2022unsupervised} also demonstrated that terrain classification from proprioceptive readings could be performed unsupervised on a full-scale quadruped and showed that this information could be used as an additional signal to improve localization. Our work differs from these in that (1) we do not use any dedicated sensor in the foot but predict the terrain properties using only standard sensors of the robot's ego-motion, and (2) we directly predict the terrain properties instead of a proxy score, which allows us to compute the cost function in simulation for multiple scenarios as in Section \ref{sec:affordance_learning}.

Another possibility is to directly predict which locomotion skill to execute from visual information~\cite{yang2023learning, loquercio2022learning}.
\citet{loquercio2022learning} learned to predict the future latent state of the policy
from a front-facing camera image to improve low-level control performance in stair climbing. An advantage of their approach is that it does not require the choice of an explicit terrain parameterization, but this comes at the cost that its visual representation is specialized to the latent of a single motor policy, so it cannot be reused for new policies or operating states, and predicting the next latent is only meaningful for egocentric images, so it cannot be used 
for novel viewpoints, as in drone-quadruped teaming or planning from satellite imagery. \citet{yang2023learning} trained a semantic visual perception module for legged quadrupeds using human demonstrations.
The resulting system imitated an operator's response to different terrains, controlling velocity and gait. This relies on a human operator to predict the terrain properties during the demonstration. Other work has learned general navigation through supervised learning on diverse robotic platforms, including legged robots~\cite{levine2023learning, shah2023gnm, shah2023vint}.
Training an omni-policy for all robots and environments enables interesting zero-shot generalization,
\begin{wrapfigure}{r}{0.5\textwidth}
    \renewcommand{\arraystretch}{1.5} 
    {\scriptsize
    \begin{tabularx}{0.38\textwidth}{|>{\hsize=1.5\hsize}X|>{\hsize=0.8\hsize}X|>{\hsize=0.8\hsize}X|>{\hsize=0.8\hsize}X|}
        \hline
        \textbf{Estimation Mode} & \textbf{Friction Loss} & \textbf{Rough Loss} & \textbf{Torque Penalty} \\ 
        \hline
        \texttt{Passive} & $1.00$ & $1.00$ & $-0.34$ \\
        \texttt{Friction} & $0.47$ & $1.06$ & $-0.87$ \\
        \texttt{Roughness} & $0.99$ & $0.72$ & $-0.84$ \\
        \texttt{Joint Fr.+Ro.} & $0.49$ & $0.80$ & $-1.18$ \\
        \hline
    \end{tabularx}
    }
    \hfill
    \raisebox{-0.5\height}{\includegraphics[width=0.10\textwidth,trim={7.5cm 0cm 7.5cm 0cm},clip]{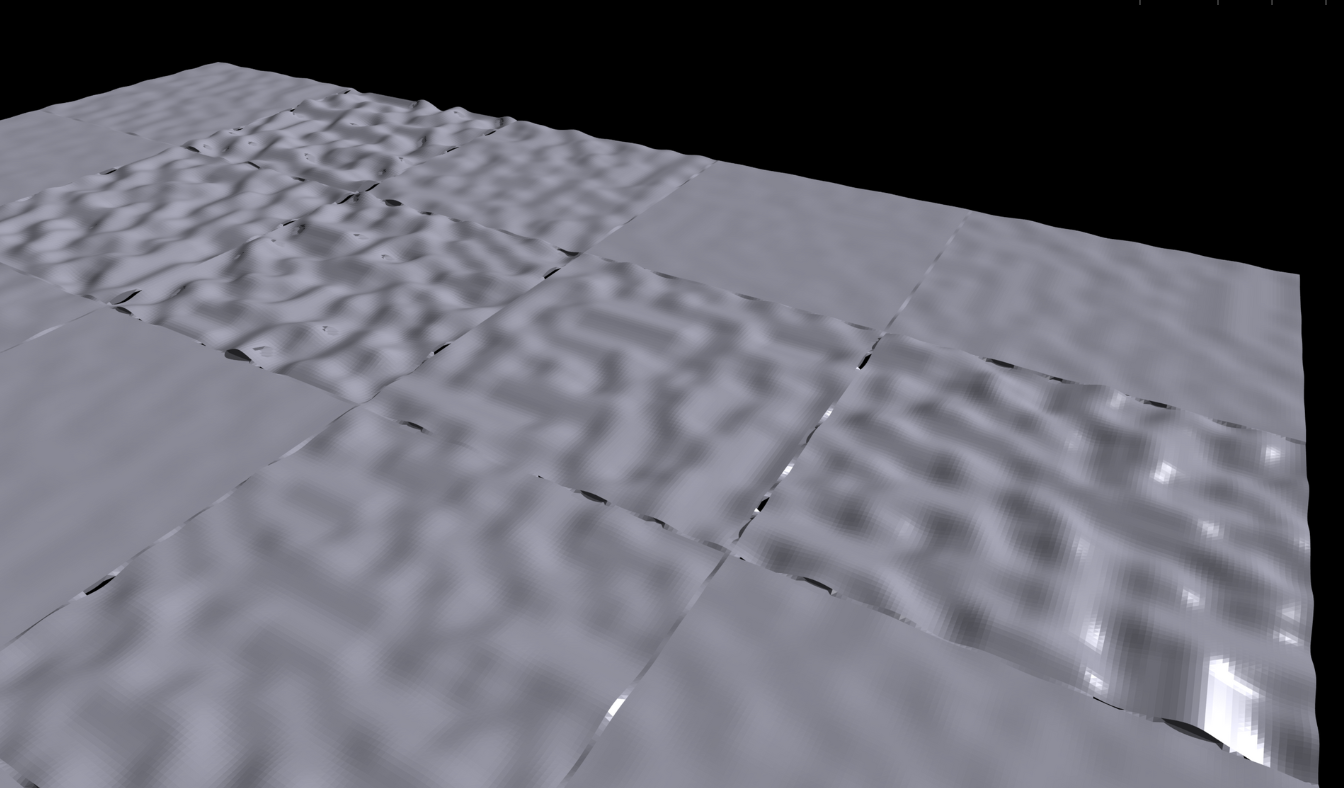}}
    \caption{\textbf{ASMP for multiple physical parameters.} Friction and roughness estimates are improved by ASMP, even when both parameters are jointly targeted. We report estimation loss for passive estimation (\texttt{None (Passive)}), active estimation of each parameter separately (\texttt{Friction}, \texttt{Roughness}), and active estimation for both parameters in a single policy (\texttt{Joint Fr.+Ro.}). Variation in torque reflects that a change in motor strategy enabled the improved estimation. 
    \textit{Right image:} Terrain with varied roughness parameter. 
    }
    \label{fig:multiple_physical_parameters}
\end{wrapfigure}
but the resulting navigation decisions do not account for the impact of embodiment (wheeled/legged) or varied operating conditions (carrying a payload) on traversability.

Several works on wheeled robots visually estimate the geometry or contact properties of the terrain through self-supervision or hand-designed criteria and then compute the traversal cost from these metrics \cite{hadsell2007online, hadsell2009learning, stavens2012self, palazzo2020domain, lee2021self, xiao2021learning, shaban2022semantic, sathyamoorthy2022terrapn, meng2023terrainnet}. Wheeled robots have a limited variety of traversal strategies compared to legged robots. Consequently, the question of selecting a locomotion controller to gather the most informative self-supervision data has not been directly addressed. Active perception suggests a solution in which a robot agent optimizes its behavior to sense the environment. This approach has been applied to vision systems \cite{bajcsy1988active, jayaraman2016look, ramakrishnan2019emergence}, and more recently has been extended to include physical interaction \cite{bohg2017interactive, van2012maximally, chu2015robotic, pathak2018zero}. 
This inspired our approach to the controller selection issue in labeling vision with proprioception for legged robots. 


\section{Discussion and Limitations}
\label{sec:conclusion}

Our work assumes a mapping between the estimated terrain properties and the robot’s performance. Friction affects the slip of the robot’s feet against the ground and the drag force of payloads and other objects, so it is an interesting factor of performance variation for practical locomotion tasks. Of course, expanding the representation of terrain physics to include additional material properties will enable more accurate performance prediction in more diverse tasks. To account for other parameters besides friction that vary in the environment, our framework can be extended to include them. For example, Figure \ref{fig:multiple_physical_parameters} shows that ASMP successfully enhances the accuracy of estimating terrain roughness in addition to friction. 

In general, ASMP may be applied for terrain labeling under two conditions: (1) a history of proprioceptive readings is sufficient to infer the parameter of interest, and (2) the parameter of interest can be effectively simulated. If these conditions are not met, a different technique besides ASMP may be necessary to collect training data. Additionally, to train our vision module, we assume terrains with different properties are visually different. If some parameters do not impact the terrain’s visual appearance, learning a vision module of the form we propose for those parameters may be impossible. Future work could explore methods to address this, such as representing uncertainty or performing an online adaptation of the estimates to the current environment based on new proprioceptive information. Finally, our labeling method assumes a mostly visible environment and does not account for occlusions. Future work can combine our system with geometric mapping to alleviate this.

The performance of generalized segmentation models is rapidly improving, and these methods will be effective for distinguishing objects relevant to robot behavior based on multimodal specifications like language descriptions or reference images~\cite{yu2023convolutions, oquab2023dinov2, zhou2017scene, cheng2022masked, kirillov2023segment}. However, the prevailing datasets for training these models do not include physical interactions, so they cannot directly predict physical properties. Moreover, the physical properties of terrain can change depending on the conditions; For example, recent rainfall may muddy a grassy field without changing its visual appearance. Therefore, it is a benefit of our method that it is informed by recently collected data from the target environment rather than relying exclusively on offline pretraining.


\clearpage
\acknowledgments{We thank the members of the Improbable AI lab for the helpful discussions and feedback on the paper. We are grateful to MIT Supercloud and the Lincoln Laboratory Supercomputing Center for providing HPC resources. This research was supported by the DARPA Machine Common Sense Program, the MIT-IBM Watson AI Lab, and the National Science Foundation under Cooperative Agreement PHY-2019786 (The NSF AI Institute for Artificial Intelligence and Fundamental Interactions, http://iaifi.org/). We acknowledge support from ONR MURI under grant number N00014-22-1-2740. This research was also sponsored by the United States Air Force Research Laboratory and the United States Air Force Artificial Intelligence Accelerator and was accomplished under Cooperative Agreement Number FA8750-19-2-1000. The views and conclusions contained in this document are those of the authors and should not be interpreted as representing the official policies, either expressed or implied, of the United States Air Force or the U.S. Government. The U.S. Government is authorized to reproduce and distribute reprints for Government purposes, notwithstanding any copyright notation herein.
}

\textbf{Author Contributions}
\begin{itemize}[leftmargin=*]
\item \textbf{Gabriel B. Margolis} ideated, implemented, and evaluated Active Sensing Motor Policies and shared ideation and implementation of the vision module and overall experimental design.
\item \textbf{Xiang Fu} shared ideation and implementation of vision module and overall experimental design.
\item \textbf{Yandong Ji} contributed ideas and supported infrastructure development during the project.
\item \textbf{Pulkit Agrawal} advised the project and contributed to its conceptual development, experimental design, positioning, and writing.
\end{itemize}


\bibliography{example_googlescholar}  

\newpage
\begin{appendix}
    \section{Reward Function for Policy Training}
    \label{sec:reward_fns}

We follow the reward structure of \cite{margolis2022walktheseways}. We remove most gait constraints but retain a fixed trotting contact schedule to facilitate sim-to-real transfer. Table \ref{tab:reward_table} lists the resulting reward terms, expressions, and weights. Table \ref{tab:definitions} summarizes our notation and lists the policy observation and action space.

\begin{figure}[t!]
\centering
\begin{minipage}[t]{0.45\textwidth}
    \centering
    \bgroup
    \def\arraystretch{1.5}
    \captionof{table}{Reward terms for \texttt{No-SE}, \texttt{Passive-SE}, and \texttt{Active-SE} policy training.}
    \label{tab:reward_table}
    \tiny
    \begin{tabular}{lrr}
    \cline{1-3}
    \multicolumn{3}{|c|}{\textbf{Reward Terms}} \\
    \cline{1-3}
    {Term}            & {Expression} & {Weight}                    \\
    \hline
    XY Vel  & \( \exp\{{-{|\textbf{v}_{xy}-\textbf{v}^{\text{cmd}}_{xy}|^2} / {\sigma_{vxy}}}\}\) & $1.0$ \\
    Yaw Vel & \(  \exp\{{-{(\boldsymbol{\omega}_z-\boldsymbol{\omega}^{\text{cmd}}_z)^2} / {\sigma_{\omega z}}}\}\) & $0.5$ \\
    Swing Phase   & $[1-\pmb{\kappa}]\mathrm{exp}\{-\delta_{\mathrm{cf}}|\mathbf{f}^{\mathrm{foot}}|^2\}  $     & $-4.0$           \\
    Stance Phase    & $ \pmb{\kappa} \mathrm{exp}\{-\delta_\mathrm{cv}|\mathbf{v}^{\mathrm{foot}}_\mathrm{xy}|^2\}$     & $-4.0$       \\
    Joint Limits & $\mathbf{1}_{q_i>q_{\mathrm{max} }||q_i<q_{\mathrm{min} }}$ & $-10.0$ \\
    Joint Torque    &     $|\boldsymbol{\tau}|^2$ & $-0.0001$        \\
    Joint Velocity & $|\dot{\mathbf{q}}|$ & $-0.0001$ \\
    Joint Acceleration & $|\ddot{\mathbf{q}}|$ & $-2.5e-7$ \\
    Hip/Thigh Collision & $\mathbf{1}_\mathrm{collision}$ & $-5.0$ \\
    Projected Gravity    &    $|\mathbf{g}_\mathrm{xy}|^2$ & $-5.0$         \\
    Action Smoothing & $|\mathbf{a}_{t-1}-\mathbf{a}_t|^2$ & $-0.1$ \\
    Action Smoothing 2 & $|\mathbf{a}_{t-2}-2\mathbf{a}_{t-1}+\mathbf{a}_{t}|^2$ & $-0.1$ \\
    \hdashline
    ASMP Bonus & $|e_t - \hat{e}_t|^2$ & $-0.3$ \\
    \hline
    \end{tabular}
    \egroup
\end{minipage}
\hfill
\begin{minipage}[t]{0.48\textwidth}
    \centering
    \bgroup
    \def\arraystretch{1.3}
    \captionof{table}{Notation, observation and action space.}
    \label{tab:definitions}
    \tiny
    \begin{tabular}{rllr}
    \toprule
                Parameter           &  Definition  &       Units & Dimension   \\
    \midrule
    \cline{1-4}
    \multicolumn{4}{|c|}{\textit{Learned Policies}} \\
    \cline{1-4}
    $\pi_\texttt{No-SE}$ & No State Est & - & - \\
    $\pi_\texttt{Passive-SE}$ & Passive State Est & - & - \\
    $\pi_\texttt{Active-SE}$ & Active State Est & - & - \\
    \hdashline
    $\pi_\texttt{Loco}$ & Policy from \cite{margolis2022walktheseways} & - & - \\
    \cline{1-4}
    \multicolumn{4}{|c|}{\textit{Policy Observation} ($\mathbf{o}$)} \\
    \cline{1-4}
    $\mathbf{q}$ & Joint Angles & \SI{}{\radian} & $12$ \\
    $\dot{\mathbf{q}}$ & Joint Velocities & \SI{}{\radian/\second} & $12$ \\
    $\mathbf{g}$ & Norm Gravity, Body Frame & \SI{}{\meter/\second^2} & $3$ \\
    $\pmb{\psi}_t$ & Body Yaw, Global Frame & \SI{}{\radian} & $1$ \\
    $\mathbf{a}_{t-1}$ & Previous Action & - & $12$ \\
    $\pmb{\theta}^\mathrm{cmd}$ & Timing Reference~\cite{margolis2022walktheseways} & - & $4$ \\
    \hdashline
    $\hat{e}_t$ & Estimator Output & - & $5$ \\
    \cline{1-4}
    \multicolumn{4}{|c|}{\textit{Policy Action} ($\mathbf{a}$)} \\
    \cline{1-4}
    $\mathbf{q}_{\text{des}}$ & Joint Position Targets & \SI{}{\radian} & $12$ \\
    \cline{1-4}
    \multicolumn{4}{|c|}{\textit{Other Quantities} }\\
    \cline{1-4}
    $\mathbf{f}^{\mathrm{foot}}$ & Foot Vertical Force & \SI{}{\newton} & $1$ \\
    $\mathbf{v}^{\mathrm{foot}}_\mathrm{xy}$ & Foot xy-Velocity & \SI{}{\meter/\second} & $1$ \\
    $\pmb{\kappa}$ & Desired Contact State & - & $1$ \\
    $\textbf{v}^{\text{cmd}}_{xy}$ & Target x-y Linear Velocity & \SI{}{\meter/\second} & $2$ \\
    $\textbf{v}_{xy}$ & Actual x-y Linear Velocity & \SI{}{\meter/\second} & $2$ \\
    $\boldsymbol{\omega}^{\text{cmd}}_z$ & Target Yaw Velocity & \SI{}{\radian/\second} & $1$ \\
    $\boldsymbol{\omega}_z$ & Actual Yaw Velocity & \SI{}{\radian/\second} & $1$ \\
    $\boldsymbol{\tau}$ & Joint Torque & \SI{}{\newton \meter} & $1$ \\
    \bottomrule
    \end{tabular}
    \egroup
\end{minipage}
\end{figure}

    \section{Hyperparameters and Architecture for Vision Module}
    \label{sec:vision_hparams}

    The vision module is structured as follows: first, we run the pretrained convolutional backbone~\cite{yu2023convolutions} on the color image to compute a feature $l_t$ for each pixel. For those patches that have an associated terrain property label $e_t$ from the proprioceptive traversal, we form a tuple ($l_t$, $e_t$). We discretize the continuous $e_t$ into bins. Finally, we train a linear model with Softmax activation to predict the bin associated with each pixel feature. Training parameters are given in Table \ref{tbl:vision_hparams}.

    \section{Simulated Evaluation}
    \label{sec:sim_eval}
\begin{figure}
\centering
    \begin{subfigure}[c]{.65\textwidth}
        \centering
        \includegraphics[width=\linewidth,trim={2cm 2cm 2cm 2cm},clip]{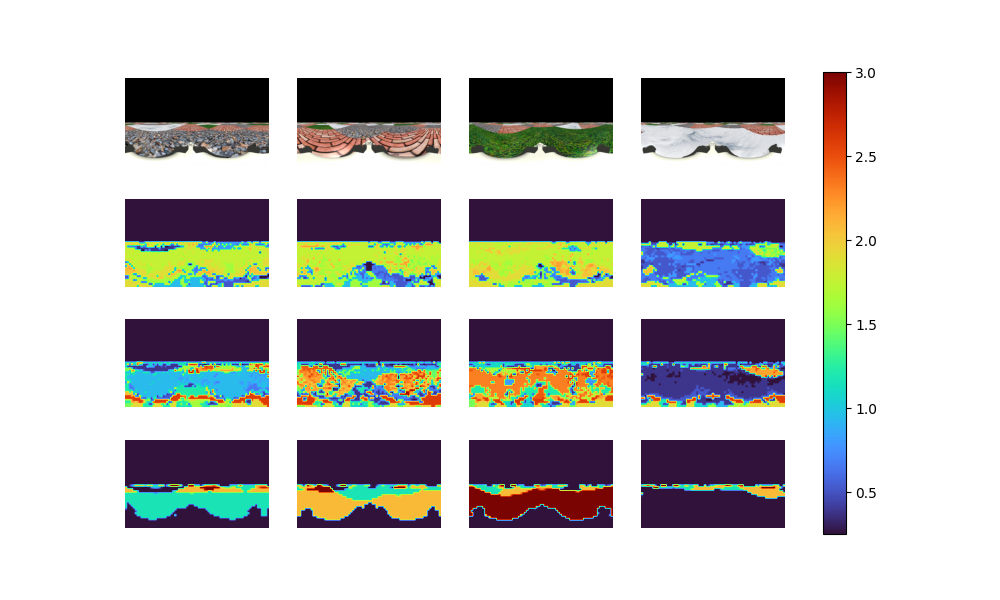} 
        \caption{Four example frames (top) and predictions (second, third row) from the simulated equirectangular camera. The model trained with passive proprioceptive sensing (second row) does not distinguish terrains with higher friction. The model trained with active proprioceptive sensing (third row) more closely matches the ground truth (bottom row).
        }
        \label{fig:sub1}
    \end{subfigure}%
    \quad
    \begin{subfigure}[c]{.32\textwidth}
        \centering
        
        \def\arraystretch{1.5}
        {
        \begin{tabular}{cc}
            \bottomrule
            \texttt{Passive-SE} & \texttt{Active-SE}\\ 
            \bottomrule
            1.23 & 0.94   \\
            \bottomrule
        \end{tabular}
        }
        \caption{RMSE for visual friction prediction across five minutes of simulated test data. Active Sensing Motor Policies enable more accurate perception.
        }
        \label{fig:sub2}
    \end{subfigure}
    \caption{\textbf{Friction inference from color images in simulation.} We collect one minute of simulated data from policies trained with and without active state estimation and compare the resulting visual inference result against the ground truth. }
    \label{fig:simvisionresult}
\end{figure}

    We collect five minutes of simulated data and train a vision module on ice, gravel, brick, and grass, assigning them arbitrary friction coefficients of $\mu=\{0.25, 1.17, 2.08, 3.0\}$ respectively. Qualitatively, the vision module learned from passive data learns to see ice but fails to distinguish between higher-friction terrains (gravel, brick, and grass). This makes sense as Figure \ref{fig:affordance_meas} shows that frictions in this range have less influence on locomotion performance. In contrast, the vision module trained on data from our Active Sensing Motor Policy learns to distinguish all four terrains. Quantitatively, ASMP results in lower dense prediction loss on images from a held-out test trajectory (Figure \ref{fig:simvisionresult}, Appendix).

    \begin{figure}
    \centering
    \includegraphics[width=0.32\linewidth,trim={0cm 0cm 0cm 12cm},clip]{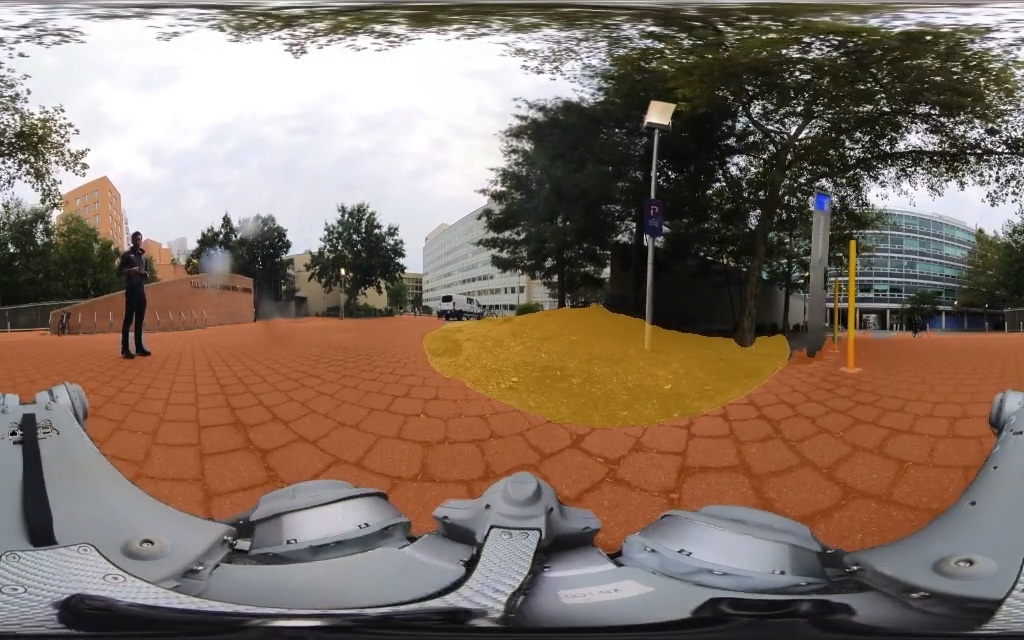}
    \includegraphics[width=0.32\linewidth,trim={1cm 1.5cm 1cm 11cm},clip]{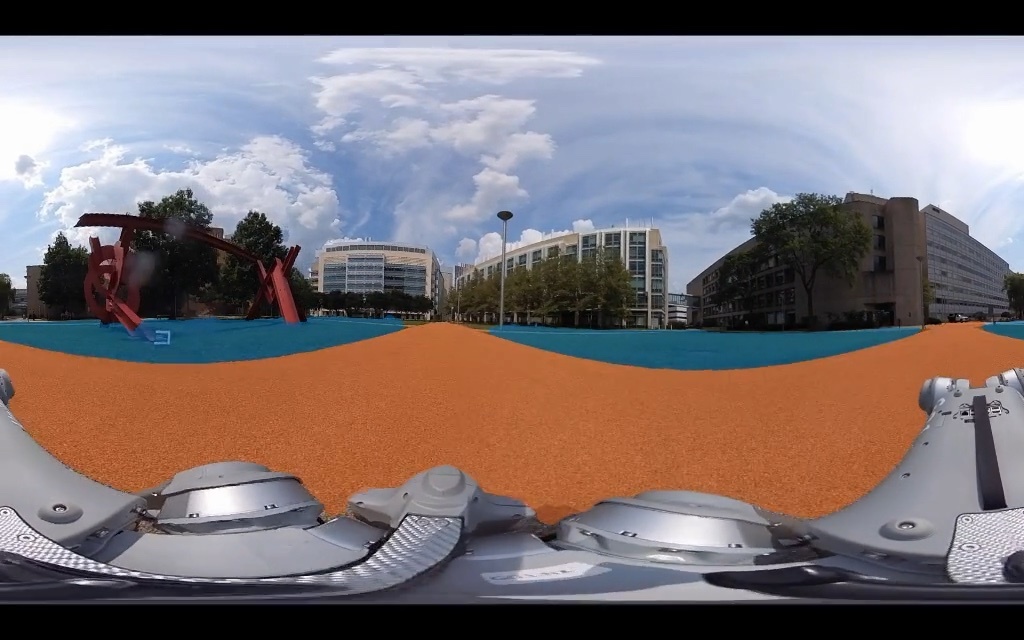}
    \includegraphics[width=0.32\linewidth,trim={0cm 0cm 0cm 12cm},clip]{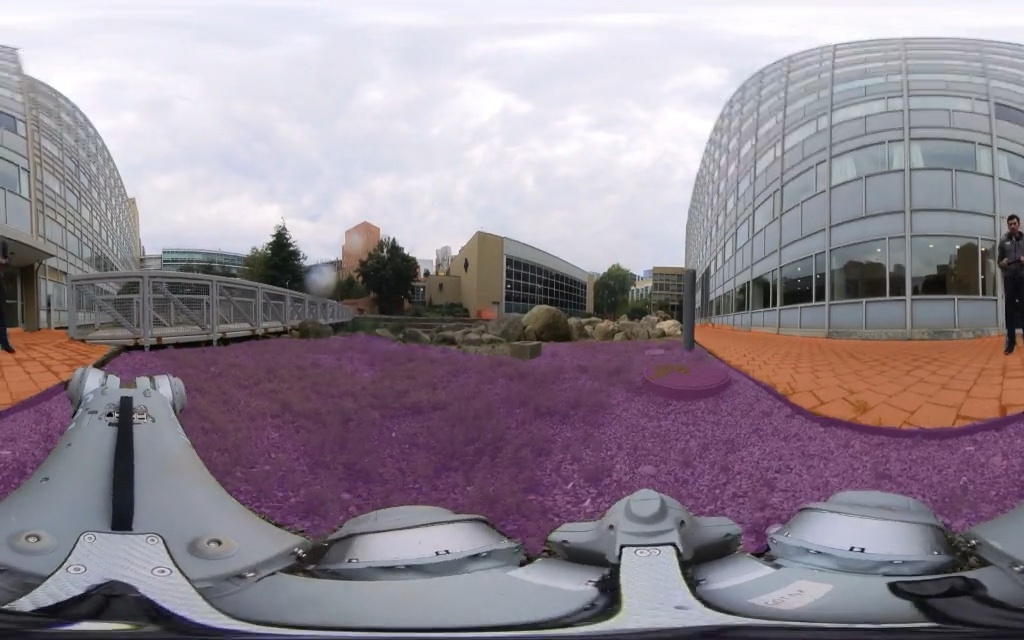}
    \caption{Example hand-labeled segmentation maps used for real-world performance analysis. Orange=pavement, yellow=dirt, blue=grass, purple=gravel.}
    \label{fig:segmentation_example}
    \end{figure}

    \section{Path Planning Procedure}
    \label{sec:path_planning}

    We use the A$^*$ algorithm~\cite{hart1968formal} to compute cost-minimal paths.
    
\begin{figure}
    \begin{minipage}[t]{.45\linewidth}
        \centering
        \raisebox{-0.0\height}{
        \small
        \begin{tabular}{rl}
        \toprule
        Hyperparameter & Value \\ [0.5ex]
        \midrule
         discount factor & $0.99$ \\ [0.5ex]
         GAE parameter & $0.95$ \\ [0.5ex]
          \# timesteps per rollout & $21$ \\ [0.5ex]
          \# epochs per rollout & $5$ \\ [0.5ex]
          \# minibatches per epoch & $4$ \\ [0.5ex]
          entropy bonus ($\alpha_2$) & $0.01$ \\ [0.5ex]
          value loss coefficient ($\alpha_1$) & $1.0$ \\ [0.5ex]
          clip range & $0.2$ \\ [0.5ex]
          reward normalization & yes \\ [0.5ex]
          learning rate & $1e-3$ \\ [0.5ex]
          \# environments & $4096$ \\ [0.5ex]
          \# total timesteps & $2.58$B \\ [0.5ex]
          optimizer & Adam \\ [0.5ex]
         \bottomrule
        \end{tabular}
        }
        \vspace{0.45cm}
        \captionof{table}
          {%
            PPO hyperparameters. 
        \label{tbl:ppo_hparams}
          }
  \end{minipage}\hfill
  \begin{minipage}[t]{.45\linewidth}
        \centering
        \raisebox{-0.0\height}{
        \small
        \begin{tabular}{rl}
        \toprule
        Hyperparameter & Value \\ [0.5ex]
        \midrule
          framerate & $5$ fps \\ [0.5ex]
          learning rate & $1e-3$ \\ [0.5ex]
          batch size & $64$ \\ [0.5ex]
          \# epochs & $20$ \\ [0.5ex]
          optimizer & Adam \\ [0.5ex]
          layers & $1$ \\ [0.5ex]
          activation & Softmax \\ [0.5ex]
          \# discrete categories & $20$ \\ [0.5ex]
         \bottomrule
        \end{tabular}
        }
        \vspace{0.45cm}
        \captionof{table}
          {%
            Vision module training hyperparameters. 
        \label{tbl:vision_hparams}
          }
  \end{minipage}
\end{figure}

\begin{table}
\centering
\caption{Numerical result of real-world friction prediction evaluation (Table \ref{fig:friction_eval}).}
\label{tab:friction_data}
\scriptsize
\begin{tabular}{|l|c|c|c|c|c|}
\hline
\textbf{Surface} & \texttt{Measured} & \texttt{ASMP (Ours)} & \texttt{Passive (Baseline)} & \texttt{Vision (Train)} & \texttt{Vision (Test)} \\
\hline
Grass & $1.45$ & $1.92 \pm 0.28$ & $2.58 \pm 0.14$ & $1.89 \pm 0.18$ & $1.80 \pm 0.26$ \\
Pavement & $0.89$ & $1.35 \pm 0.29$ & $2.42 \pm 0.13$ & $1.32 \pm 0.22$ & $1.35 \pm 0.22$ \\
Dirt & $0.63$ & $0.90 \pm 0.23$ & $2.12 \pm 0.26$ & $0.90 \pm 0.23$ & $1.09 \pm 0.36$ \\
Gravel & $0.74$ & $0.90 \pm 0.39$ & $1.75 \pm 0.47$ & $0.81 \pm 0.26$ & $1.12 \pm 0.39$ \\
\hline

\end{tabular}
\end{table}

\end{appendix}

\end{document}